\documentclass[11pt, a4paper, logo, copyright]{gensyn}

\usepackage{algorithm}
\usepackage{algorithmic}
\usepackage{bm}
\usepackage{multirow}

\theoremstyle{remark}

\newcommand{\RETURN}{\STATE \textbf{return}~}
\newcommand{\method}{DEI}

\newcommand{\qdScore}{\text{QD-Score}}
\newcommand{\bc}{\mathbf{bc}}
\newcommand{\cA}{\mathcal{A}}
\newcommand{\cB}{\mathcal{B}}
\newcommand{\cO}{\mathcal{O}}

\newcommand{\cR}{\mathcal{R}}
\newcommand{\cH}{\mathcal{H}}
\newcommand{\cT}{\mathcal{T}}

\renewcommand{\emph}[1]{\textit{#1}}
\title{DEI: Diversity in Evolutionary Inference for Quality-Diversity Search}
\author{John Donaghy}
\author{Shikhar Rastogi}
\affil{Gensyn}
\correspondingauthor={John Donaghy \texttt{<johnny@gensyn.ai>}}

\begin{document}

\begin{abstract}
We present \method{}: Diversity in Evolutionary Inference,
a distributed Quality-Diversity (QD) search framework that assigns
heterogeneous large language models (LLMs) as mutation operators across peer nodes
communicating with non-blocking collective operations. Unlike homogeneous parallel search, which
replicates a single model's inductive biases across all workers, \method{} treats
each LLM's distinct creative prior as a complementary source of behavioral novelty.
Extending the Digital Red Queen framework with \method{}, nodes share local optimal solutions
at the end of each round to seed the next round's population.  This creates
cross-model adversarial pressure that drives robustness beyond intra-model self-play.
Evaluated on the Core War domain, a competitive programming benchmark in which
Redcode warrior programs battle inside a simulated machine, a four-node heterogeneous
ensemble (GPT-5.4-mini, Claude Sonnet 4.6, GPT-5.2, and Claude Haiku 4.5) achieves
$+124\%$ higher merged-archive QD-Score ($45.90$ vs.\ $20.46$) and $+28\%$ higher
coverage ($80.6\%$ vs.\ $63.0\%$ of cells) than a single-node baseline at equal
total LLM-call budget.  The heterogeneous ensemble also outperforms an equally-budgeted homogeneous
ensemble on QD-Score, coverage, and held-out solution generality across all four
model families. These results provide the first
empirical evidence that \emph{model diversity}, not merely parallelism, is the key
driver of gain in distributed LLM-based QD search.
\end{abstract}

\maketitle

\section{Introduction}
\label{sec:introduction}

Evolutionary algorithms search for high-performing solutions by maintaining a
population of candidates and selecting the best from the population. 
The population is iteratively updated by generating new candidates either from scratch or 
by mutating existing solutions.
Mutation is performed by the \emph{mutation operator} which is a general procedure
for taking an existing solution and perturbing it to yield a variant, whose
fitness is then evaluated for inclusion in the population.

Large language models have emerged as surprisingly effective creative operators in
evolutionary search.  They generate novel program variants, mutate existing solutions
with semantic awareness, and reason about fitness landscapes in ways that classic
genetic operators cannot. Yet every LLM encodes a particular distribution over
programs shaped by its training data, architecture, and alignment procedure. A model
trained predominantly on Python tutorials will favor different control-flow patterns
than one trained on competitive-programming corpora.  A model fine-tuned for
instruction following will hedge in ways that a code-completion model will not.
These inductive biases should be harnessed deliberately to cover
more of a behavioral space than any single model can reach alone.

Existing approaches to parallel LLM search treat scaling as a matter of
\emph{computation}, not \emph{cognition}. FunSearch~\citep{romeraparedes2023funsearch}
runs many independent LLM calls in parallel, but all calls go to the same model.
Diversity emerges only from stochastic sampling, not from fundamentally different
generative priors. The Digital Red Queen (DRQ) framework~\citep{kumar2026digitalredqueen}
introduced adversarial evolutionary pressure into MAP-Elites~\citep{mouret2015mapelites}
Quality Diversity (QD) search by using round champions as opponents, but in its original formulation a
single node runs a single model. Blind spots in that model's generative distribution
become permanent gaps in the archive.

We argue for a qualitatively different regime: \textit{parallel cognition}.
When diverse LLMs simultaneously explore a shared behavioral
space, each model's inductive bias causes it to preferentially populate distinct
regions of the Quality-Diversity archive~\citep{chatzilygeroudis2020qdsurvey}.
A growing body of work in reinforcement learning and multi-agent reasoning shows that
explicitly promoting generative diversity, whether through diversity-aware RL
objectives~\citep{li2025darling, hu2025diver} or multi-agent debate across several
LLMs~\citep{liang2024encouraging}, improves exploration and downstream task
performance. We apply the same intuition to distributed Quality-Diversity search.
Champions that one model considers elite may inhabit niches that another model never
generates, providing ready-made seeds for cross-pollination. When these
champions are shared as opponents via non-blocking communications, every node may
progress at its own pace while competing against strategies it would never have generated
internally, driving robustness that pure intra-model self-play cannot provide.

\begin{figure}[t]
\centering
\includegraphics[width=\columnwidth]{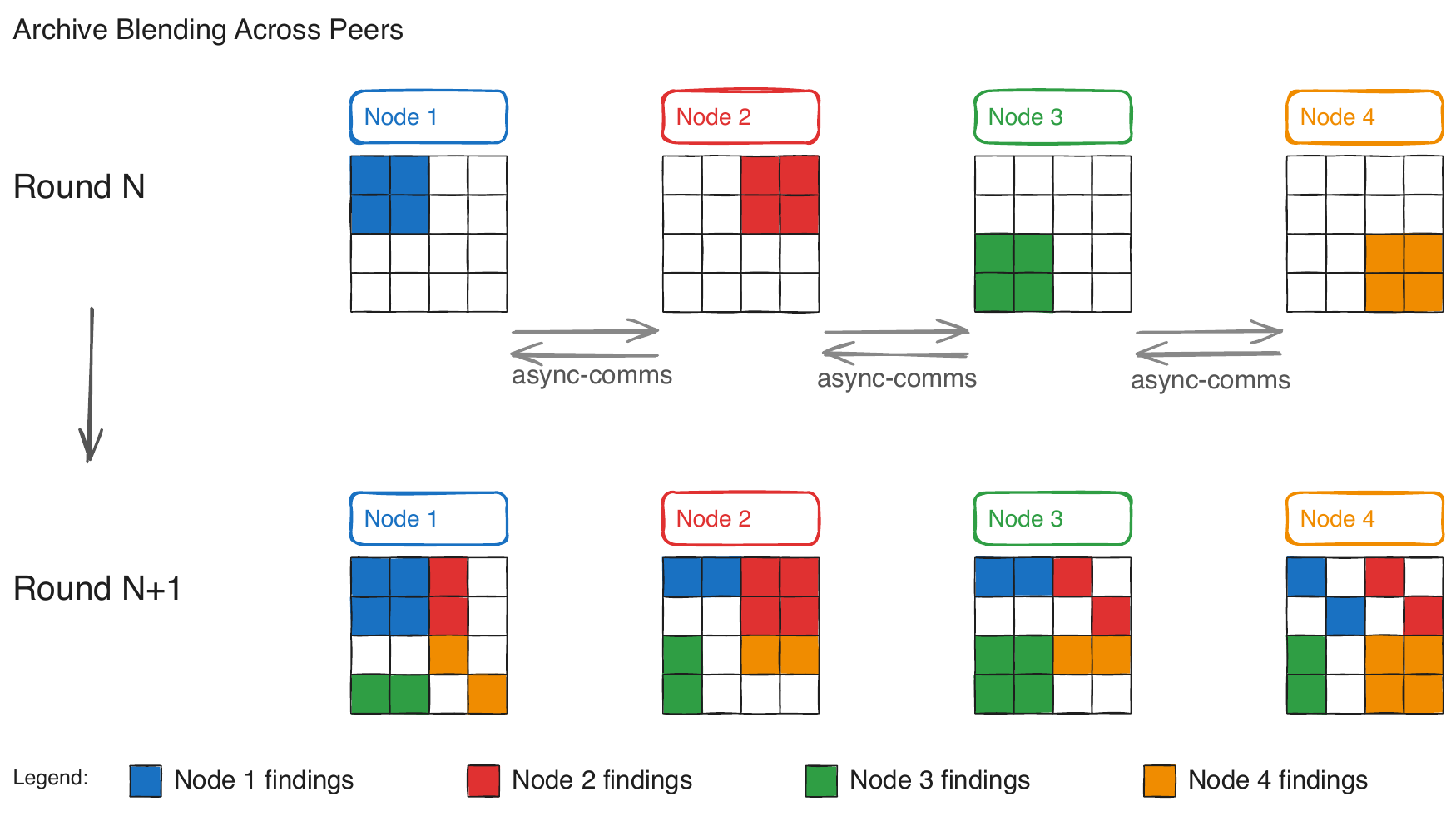}
\caption{Heterogeneous ensembles as parallel cognition. In an early round, each
node's inductive bias preferentially populates a distinct region of the shared
Quality-Diversity archive (top). Asynchronous champion sharing then seeds peers
with elites from niches they would rarely generate internally, so archives begin
to blend across nodes in subsequent rounds (bottom).}
\label{fig:archive_blending}
\end{figure}

Realizing this vision requires solving a practical systems problem: heterogeneous LLM
nodes operate at wildly different latencies. A local Qwen-35B instance running on
Apple Silicon with MLX may take ten seconds per call; a frontier model may
respond in under two. A synchronization barrier would throttle the entire ensemble
to the slowest participant. We therefore develop fully asynchronous collective communications,
allowing each node advance on independently, regardless of latencies.

\paragraph{The main contributions of the paper can be listed as:}
\begin{itemize}
  \item \textbf{DEI:} A distributed QD search framework that
    explicitly exploits \emph{heterogeneous} LLM ensembles as mutation operators,
    assigning each model a distinct node in the search graph.
  \item \textbf{Async gossip integration:} A fully asynchronous champion-sharing
    protocol that enables heterogeneous hardware
    participation without synchronization barriers achieving higher throughput than a synchronous approach.
  \item \textbf{Diversity aware improvements on generality, archive coverage and QD-Score:} By holding total compute
    budget fixed across conditions, we show the gain from heterogeneous nodes over
    homogeneous nodes cannot be attributed to extra computation, isolating diversity
    as the causal factor.
\end{itemize}

\section{Background}
\label{sec:background}
We use the procedure developed by \citet{kumar2026digitalredqueen} as a single node baseline 
for extension to evolution with a diverse ensemble of mutation operators.

\subsection{Core War and the MARS Simulator}
\label{sec:bg:corewar}

Core War is a competitive programming game in which two or more \emph{warrior}
programs, written in the Redcode assembly language, battle inside a virtual circular
memory called the Memory Array Redcode Simulator (MARS). Each warrior attempts to
crash all opponents by overwriting their instruction pointers or halting their
processes, while surviving as long as possible itself. The memory array wraps
around, all addresses are relative and modular, and multiple processes per warrior
are permitted, making the space of viable strategies remarkably rich: imps that
replicate forward, dwarfs that bomb the core with \texttt{DAT} instructions,
scanners that locate and attack enemy code, and fortress warriors that protect a
compact core with a paper layer.

In our experiments, the MARS configuration duplicates the hyperparameters used by the single
node implementation in the original DRQ paper~\citet{kumar2026digitalredqueen}. See Appendix~\ref{app:mars}
for the details.  The fitness of a warrior $w_i$ against a
set of opponents $\cO$ is:
\begin{equation}
  f(w_i, \cO) = \sum_{\tau \in \cT} \frac{N}{\cT} \frac{A^i_\tau}{\sum_{o \in \cO} A^o_\tau}
  \label{eq:fitness}
\end{equation}
where $\cT$ is the number of simulation timesteps, and $A^i_\tau$ indicates if warrior
 i is alive at timestep $\tau$, and $N$ is the number of warriors in the battle.  

\subsection{MAP-Elites and Quality-Diversity Search}
\label{sec:bg:mapelites}

Quality-Diversity (QD) search~\citep{chatzilygeroudis2020qdsurvey} seeks not a
single optimal solution but a large, diverse collection of high-performing solutions
spanning a user-defined behavioral space. MAP-Elites~\citep{mouret2015mapelites} is
the canonical QD algorithm: it maintains an \emph{archive} $\cA$ of cells indexed
by a discretized behavioral characteristic (BC) space $\cB$. Each cell stores at
most one elite, the highest-fitness solution observed at that BC coordinate. At each
iteration, an existing elite is selected, mutated, evaluated, and placed in the
archive if it either fills an empty cell or improves over the current occupant.

Keeping with~\citet{kumar2026digitalredqueen}, we use a two-dimensional BC space defined by:
\begin{itemize}
  \item \textbf{Time-Space Product (TSP):} the product of the warrior's code length
    and its average lifespan across battles, capturing the trade-off between spatial
    footprint and temporal persistence.
  \item \textbf{Memory Coverage (MC):} the fraction of the core that the warrior
    touches (reads, writes, or executes) during a battle, reflecting how aggressively
    it explores the address space.
\end{itemize}

Differing from the fitness used for archive placement (Equation~\eqref{eq:fitness}) which varies with iteration,
we evaluate solution quality using a static held out cohort of human authored warriors.
Here we take a solution's Generality as indicative of its quality:
\begin{equation}
  \text{generality}(w_r, \cH) = \frac{|\{h \in \cH : \texttt{win\_tie}(w_r,h)\}|}{|\cH|}
  \label{eq:generality}
\end{equation}
where $r$ is the round number, $\cH$ is a fixed held-out set of human-authored
warriors, and $\texttt{win\_tie}()$ is 1 if warrior $w_r$ wins/ties against warrior $h$ and 0 if it loses.

\subsection{Digital Red Queen (DRQ)}
\label{sec:bg:drq}

The Digital Red Queen framework~\citep{kumar2026digitalredqueen} integrates an LLM
as the primary mutation operator inside MAP-Elites, replacing hand-crafted genetic
operators with natural-language-conditioned program generation. At each iteration,
an elite is randomly sampled from the archive, are serialized into a prompt and 
mutated by the LLM, or a new warrior is generated from scratch by the LLM.
The adversarial Red Queen pressure is introduced by maintaining a pool of \emph{round champions}, 
the highest-fitness warrior from each completed round, and using them as the opponent set
in Equation~\eqref{eq:fitness}. As the archive improves, the opponent pool grows
stronger, continuously raising the selection pressure in a co-evolutionary arms race.

\subsection{Asynchronous Champion Sharing}
\label{sec:bg:gossip}

In a distributed evolutionary system with heterogeneous LLMs, nodes operate at
different latencies, a local open-weight model may take an order of magnitude
longer per call than a frontier model. A synchronization barrier would throttle the
entire ensemble to the slowest participant. We therefore adopt a fully
asynchronous all-gather communication pattern: each node publishes its round champion
to its peers whenever ready, and peers consume incoming champions at their
own pace. This design achieves reliable champion propagation without
synchronization barriers, and is detailed in Appendix~\ref{app:implementation}.

\section{Method}
\label{sec:method}

\subsection{System Architecture}

Each participating node in \method{} instantiates two cooperating subsystems:
(i) an \emph{asynchronous messaging layer} that handles champion sharing with
peers, and (ii) a local \emph{DRQ optimizer} that couples MAP-Elites with an
LLM mutation engine.  The DRQ code is the open source implementation supplied by~\citet{kumar2026digitalredqueen},
and was used unmodified for the \emph{solo} runs with the communication layer included for the ensemble runs.

The communication layer handles asynchronous champion propagation between nodes. The DRQ optimizer runs locally and independently on each node,
querying its assigned LLM for warrior generation and mutation.

\subsection{Per-Node DRQ Loop}

Within each round a node performs $T$ LLM calls. With probability 0.1 the LLM
\emph{generates} a new warrior from scratch; with probability 0.9 it \emph{mutates}
an existing warrior sampled uniformly from the current archive. The fitness of a
candidate warrior $w$ under a round $r$'s opponent pool $\cO_r$ is given by Equation~\eqref{eq:fitness}.
The pool $\cO_r$ comprises the round's initial seed warriors plus champion warriors retained
from the previous K rounds, and the warriors collected
from peer nodes.

At the end of each round the \emph{champion} is selected:
\begin{equation}
  \hat{w}_r = \operatorname*{arg\,max}_{w \in \cA_r} f(w, \cO_r).
\end{equation}
  The champion is broadcast to the node's peers. The node
then drains its receive buffer to collect champions published by peer nodes during the
same round. Received champions are (a) appended to the local opponent pool, providing
cross-model adversarial pressure, and (b) seeded into the local archive if they
occupy previously empty cells. The complete procedure is formalized in
Algorithm~\ref{alg:collab-drq}.

\begin{algorithm}[t]
\caption{DEI (node $i$)}
\label{alg:collab-drq}
\begin{algorithmic}[1]
\REQUIRE $\text{LLM}_i$, initial opponents $\cO_0$, rounds $R$, iters $T$
\STATE $\cA_i \leftarrow \{\}$; \quad $\cO_i \leftarrow \cO_0$
\FOR{$r = 1$ \textbf{to} $R$}
  \FOR{$t = 1$ \textbf{to} $T$}
    \IF{$\text{rand}() < 0.1$}
      \STATE $w \leftarrow \text{LLM}_i.\textsc{Generate}(\text{new\_prompt})$
    \ELSE
      \STATE $p \leftarrow \textsc{Sample}(\cA_i)$
      \STATE $w \leftarrow \text{LLM}_i.\textsc{Mutate}(\text{mutate\_prompt}, p)$
    \ENDIF
    \STATE $\bc \leftarrow \textsc{ComputeBC}(w)$ \COMMENT{$(TSP, MC)$ from simulation}
    \STATE $f \leftarrow \textsc{Evaluate}(w, \cO_i)$ \COMMENT{Eq.~\eqref{eq:fitness}}
    \STATE $\cA_i \leftarrow \textsc{MapElitesUpdate}(\cA_i, \bc, w, f)$
  \ENDFOR
  \STATE $\hat{w}_r \leftarrow \operatorname*{arg\,max}_{w \in \cA_i} f(w, \cO_i)$
  \STATE $\textsc{Publish}(\hat{w}_r)$
  \STATE $\cR \leftarrow \textsc{Drain}()$
  \STATE $\cO_i \leftarrow \cO_i \cup \cR$
  \STATE $\cA_i \leftarrow \textsc{Seed}(\cA_i, \cR)$
\ENDFOR
\RETURN $\operatorname*{arg\,max}_{w \in \cA_i} f(w, \cO_i)$
\end{algorithmic}
\end{algorithm}

\subsection{Experimental Conditions}

To disentangle the effect of additional compute from the effect of model diversity,
we design three conditions, all holding the total LLM call budget constant.
Table~\ref{tab:conditions} summarizes the design. 

\begin{table}[t]
\centering
\caption{Experimental conditions. The budget per node is inversely proportional to $N$, keeping total LLM calls fixed.}
\label{tab:conditions}
\resizebox{\columnwidth}{!}{%
\begin{tabular}{lccl}
\toprule
\textbf{Condition} & $N$ & \textbf{Diversity} & \textbf{Purpose} \\
\midrule
Solo [DRQ]         & 1 & n/a           & Single-node baseline (each LLM) \\
Homogeneous Ensemble & 4 & Homogeneous   & Same LLM on all nodes \\
Diverse Ensemble     & 4 & Heterogeneous & Mixed LLMs \\
\bottomrule
\end{tabular}
}%
\end{table}

The heterogeneous condition (Diverse Ensemble) distributes nodes across four models:
GPT-5.4-mini, Claude Sonnet~4.6, GPT-5.2, and Claude Haiku~4.5. The homogeneous condition
runs all four nodes with the same model, giving four Homogeneous Ensemble variants.  The solo
condition runs the DRQ algorithm on a single node in isolation.

\subsection{Evaluation Metrics}

We measure performance across four dimensions:

\begin{description}
  \item[Generality:] Fraction of a held-out set of human authored warriors the final champion
    defeats or ties Equation~\eqref{eq:generality}.
  \item[Archive coverage:] Fraction of MAP-Elites grid cells occupied:
    \begin{equation*}
      \text{coverage}(\cA) = \frac{|\{b \in \cB : \cA(b) \neq \emptyset\}|}{|\cB|},
    \end{equation*}
    where $\cB$ is the full set of discretized behavioral cells and $\cA(b)$ denotes
    the elite stored at cell $b$. For distributed conditions (Diverse Ensemble,
    Homogeneous Ensemble) we report the coverage of the \emph{cross-node merged archive},
    the union of the four nodes' final-round archives, keeping the best
    warrior per cell. This equalises total LLM-call budget against Solo
    ($4 \times 62 \approx 1 \times 250$ iters per round), so coverage differences
    reflect ensemble effects rather than per-node budget asymmetry.
  \item[QD-Score:] Sum of fitness values across all occupied archive cells:
    \begin{equation*}
      \qdScore(\cA) = \sum_{b \in \cB} f(\cA(b),\, \cO)\cdot\mathbf{1}[\cA(b) \neq \emptyset],
    \end{equation*}
    where $f(\cA(b), \cO)$ is the fitness of the elite at cell $b$ against opponent
    set $\cO$ (Equation~\eqref{eq:fitness}). QD-Score rewards both breadth (coverage)
    and quality (per-cell fitness), making it sensitive to improvements on either axis.
  \item[Niche novelty $\eta$:] For each received champion $w$,
    $\eta = \mathbb{E}[\mathbf{1}[\bc(w) \notin \cA_i^{(r-1)}]]$, the fraction of
    received champions landing in previously empty cells of the receiving node's
    archive.
\end{description}

\section{Results}
\label{sec:results}

\begin{figure}[t]
\centering
\includegraphics[width=\columnwidth]{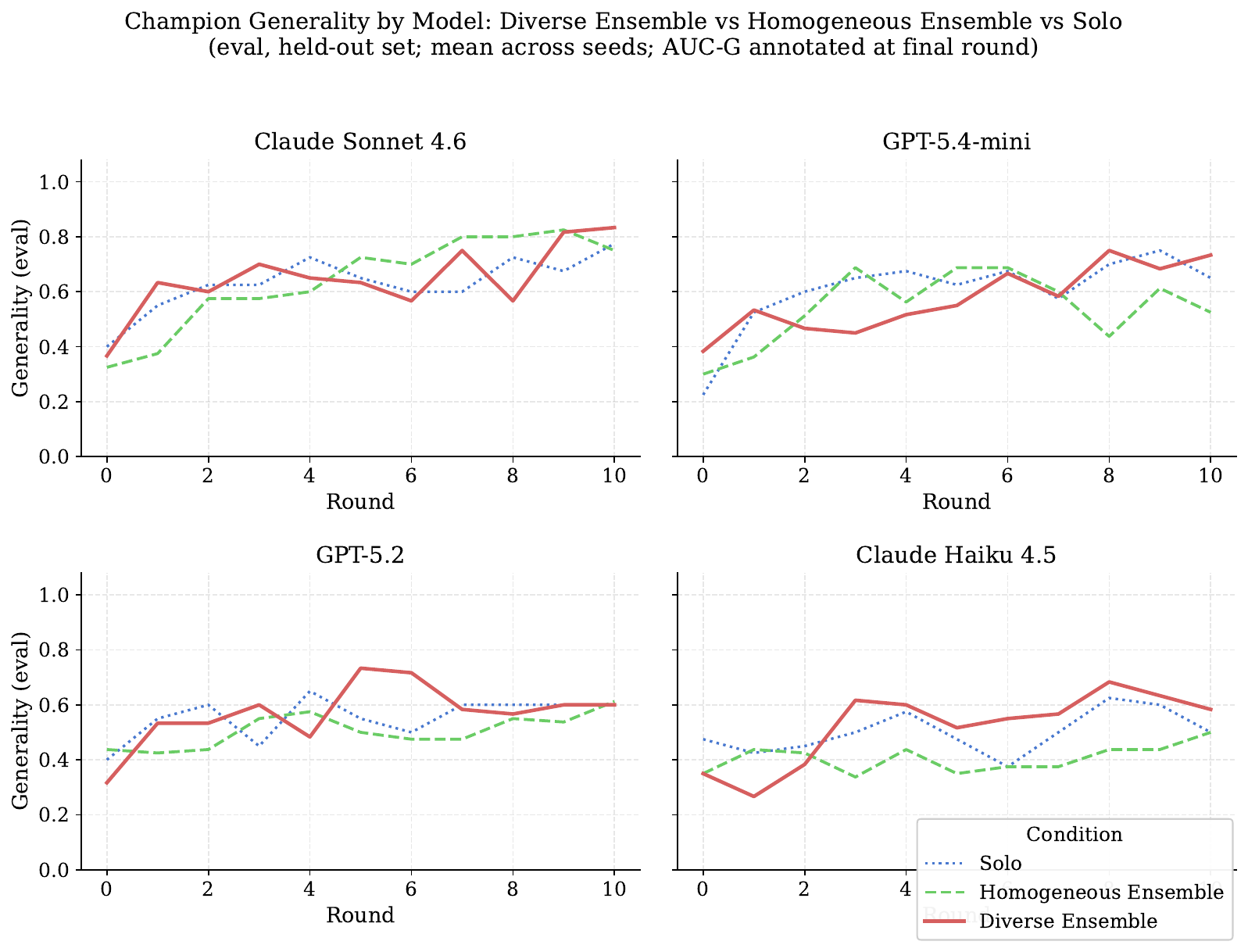}
\caption{Champion generality (eval, held-out) over rounds per model.
Each panel shows Diverse Ensemble, Homogeneous Ensemble, and Solo for one LLM.
Diverse Ensemble achieves the highest generality for all four models,
consistent with the diversity hypothesis.}
\label{fig:generality_grid}
\end{figure}

\subsection{Individual Node}
Results in Table~\ref{tab:main_results} confirm the primary hypothesis across all four model families:
Diverse collaboration achieves higher generality than homogeneous collaboration,
which in turn outperforms or is comparable with solo across most conditions, under equal total LLM call
budget.

Figure~\ref{fig:generality_grid} shows the mean champion generality over rounds, broken
down by model.  

Niche novelty, while undefined in the solo case, meaningfully improves when moving from homogeneous to diverse ensembles.
This supports the hypothesis that the breadth and quality of the archive is being increased due to collaborative peers, rather
than originating from the node's LLM directly. 

\begin{table*}[t]
\centering
\caption{Main results across all conditions. Mean $\pm$ std shown;
bold = best within each model group. Horizontal rules separate model families.}
\label{tab:main_results}
\begin{tabular}{lcc}
\toprule
\textbf{Condition} & \textbf{Peak Gen.} & \textbf{Niche Nov.\ $\eta$} \\
\midrule
Diverse Ensemble (Claude Sonnet 4.6) & \textbf{0.850} $\pm$ \textbf{0.087} & \textbf{0.483} $\pm$ \textbf{0.120} \\  
Homogeneous Ensemble (Claude Sonnet 4.6) & 0.825 $\pm$ 0.106 & 0.348 $\pm$ 0.039 \\  
Solo [DRQ] (Claude Sonnet 4.6) & 0.775 $\pm$ 0.035 & \textemdash \\  
\midrule
Diverse Ensemble (GPT-5.4-mini) & \textbf{0.767} $\pm$ \textbf{0.076} & \textbf{0.422} $\pm$ \textbf{0.072} \\  
Homogeneous Ensemble (GPT-5.4-mini) & 0.725 $\pm$ 0.029 & 0.119 $\pm$ 0.013 \\  
Solo [DRQ] (GPT-5.4-mini)    & 0.750 $\pm$ 0.000 & \textemdash \\  
\midrule
Diverse Ensemble (GPT-5.2)   & \textbf{0.767} $\pm$ \textbf{0.076} & \textbf{0.454} $\pm$ \textbf{0.070} \\  
Homogeneous Ensemble (GPT-5.2) & 0.700 $\pm$ 0.000 & 0.091 $\pm$ 0.010 \\  
Solo [DRQ] (GPT-5.2)         & 0.650 & \textemdash \\  
\midrule
Diverse Ensemble (Claude Haiku 4.5) & \textbf{0.700} $\pm$ \textbf{0.050} & \textbf{0.443} $\pm$ \textbf{0.132} \\  
Homogeneous Ensemble (Claude Haiku 4.5) & 0.538 $\pm$ 0.063 & 0.139 $\pm$ 0.040 \\  
Solo [DRQ] (Claude Haiku 4.5) & 0.650 $\pm$ 0.141 & \textemdash \\  
\bottomrule
\end{tabular}
\end{table*}

\subsection{Merged Archive}

The generality and niche novelty metrics above reflect individual node performance.
A complementary view asks: when the archives from all four collaborative nodes
are pooled into a single merged archive (keeping the best warrior per
behavioral cell), how does the combined population compare to a solo node
at equal total compute?

\begin{figure}[t]
\centering
\includegraphics[width=\columnwidth]{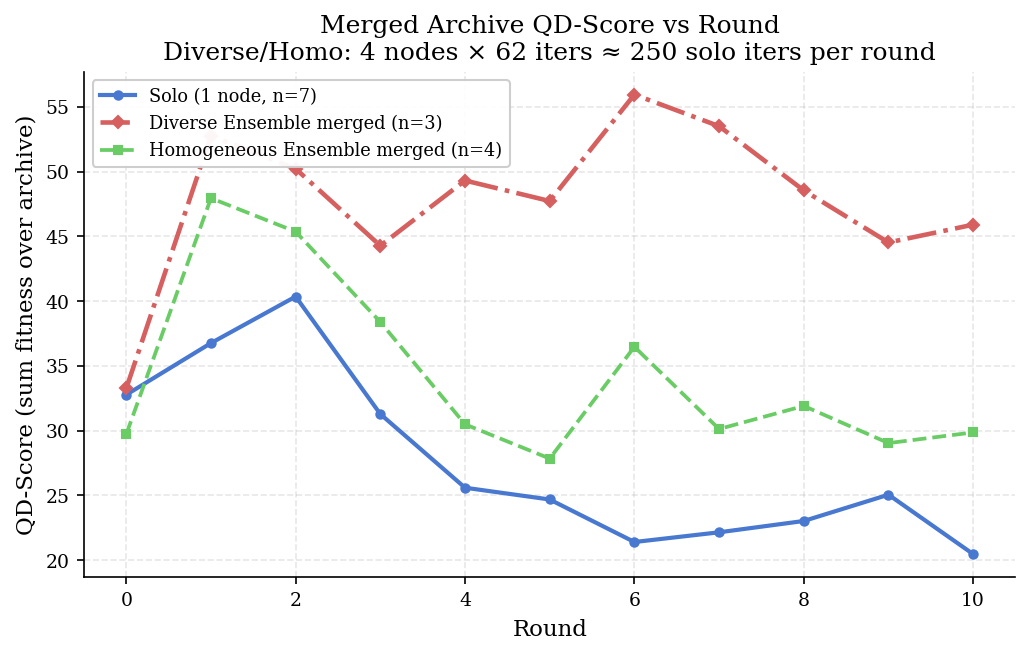}
\caption{Merged archive QD-Score vs.\ round at equal total compute
(4 nodes $\times$ 62 iters $\approx$ 250 solo iters per round).
Each curve is the mean across all available runs. Both ensembles
outperform Solo from round~1 onward; the homogeneous ensemble sustains
the highest QD-Score through the final round, while the diverse ensemble
leads on archive coverage (Table~\ref{tab:merged_archive}).}
\label{fig:merged_qdscore}
\end{figure}

Figure~\ref{fig:merged_qdscore} shows merged QD-Score vs.\ round, averaged
across all available runs.
At each round, the merged diverse archive is built from 4 nodes
$\times$ 62 iterations = 248 total LLM calls, matching the solo budget of
250 calls per round.

Table~\ref{tab:merged_archive} summarises the final-round merged archive.
The homogeneous ensemble does not outperform the solo case in coverage.
The diverse ensemble outperforms the solo baseline and the homogeneous ensemble on coverage and QD-Score, achieving $+124\%$
higher QD-Score and $+28\%$ more coverage than a solo node.
The diverse ensemble yields the broadest coverage of the cell space ($80\%$), while the homogeneous ensemble concentrates
higher per-cell fitness which helps it to outperform the solo case on QD-Score at the final round.

\begin{table}[t]
\centering
\caption{Cross-node merged archive at the final round vs.\ the single-node
Solo archive, at equal total LLM-call budget.}
\label{tab:merged_archive}
\begin{tabular}{lcc}
\toprule
\textbf{Condition} & \textbf{Coverage} & \textbf{QD-Score} \\
\midrule
Solo           & 63.0\% & 20.46 \\
Homo merged    & 59.0\% & 29.85 \\
Diverse merged & 80.6\% & 45.90 \\
\bottomrule
\end{tabular}
\end{table}

\section{Discussion}
\label{sec:discussion}
The central finding, that diverse ensembles outperform homogeneous ensembles,
which in turn outperform solo, holds across two of the three hypothesized complementary measurement
axes: individual-node generality on the held-out set,
(Table~\ref{tab:main_results}),
and the merged population archive Coverage/QD-Score (Figure~\ref{fig:merged_qdscore}).
The merged archive comparison is particularly compelling because it is
compute-equalised by construction: four collaborative nodes running 62
iterations each contribute the same total LLM calls as one solo node
running 250, yet the diverse merged archive achieves higher
Coverage and QD-Score by the final round.
These results reinforce the intuition from recent diversity-aware reinforcement
learning~\citep{li2025darling, hu2025diver} and multi-agent reasoning
work~\citep{liang2024encouraging}: explicitly sourcing generative diversity,
here by assigning distinct LLM families to distinct nodes, yields measurable
gains over single-distribution baselines at matched compute.

The fully asynchronous design means that adding a slow node (e.g., local MLX
inference on a laptop) improves coverage without penalizing faster nodes.

We study Core War as a self-contained benchmark with a clear fitness function and
well-defined behavioral characteristics. The generality of our findings to domains
with less structured BC spaces or more expensive fitness evaluations remains to be
established. While there is no theoretical reason to believe these findings would not hold on 
other domains, our emprical results are strictly within this domain, it requires further testing to verify
the applicability elsewhere. 

Promising directions for further exploration include: adaptive topology (dynamically connecting nodes whose
archives are most complementary), heterogeneous BC axes per node (each model tracks
different behavioral dimensions), and extension to multi-agent collaborative tasks beyond
Core War.

\section{Related Work}
\label{sec:related}

\paragraph{LLMs as evolutionary operators:}
The use of LLMs as mutation and crossover operators has grown rapidly since the
seminal Evolution through Large Models~\citep{lehman2022evolutionlarge} framework,
which showed that an LLM pre-trained on code can serve as a context-aware mutation
operator in an evolutionary algorithm. EvoPrompting~\citep{chen2023evoprompting}
applies LLM-driven code-level neural architecture search, and
OPRO~\citep{yang2023opro} frames optimization broadly as in-context learning.
FunSearch~\citep{romeraparedes2023funsearch} scales this to a distributed setting
with a large homogeneous pool of LLM evaluators searching for combinatorial
mathematical functions. \citet{wu2024evollmsurvey} provide a comprehensive survey
and roadmap of evolutionary computation in the LLM era.
AlphaEvolve~\citep{novikov2025alphaevolvecodingagentscientific} uses an ensemble of
LLMs within an evolutionary loop, but restricts diversity to model \emph{size}
(pairing a smaller and a larger model from the same family) to ``balance
computational throughput with the quality of generated solutions.'' In contrast,
we diversify across model \emph{families} with the explicit goal of increasing
sample diversity. We are the first to study such \emph{heterogeneous} model
ensembles in distributed evolutionary search.

\paragraph{Quality-Diversity algorithms:}
MAP-Elites~\citep{mouret2015mapelites} introduced the foundational archive-plus-BC
structure that all QD methods build on.
\citet{chatzilygeroudis2020qdsurvey} provide a thorough taxonomy.
Differentiable QD~\citep{fontaine2021differentiableqd} extends the framework to
gradient-based methods; CVT-MAP-Elites~\citep{vassiliades2016cvt} scales the
archive to high-dimensional BC spaces via centroidal Voronoi tessellations.
\citet{cully2018hierarchical} introduce unsupervised behavioral descriptors that
allow an archive to adapt its BC axes without hand-engineering, a direction
orthogonal to, but complementary with, our multi-model approach.
Quality-Diversity through AI Feedback~\citep{bradley2023qdaif} replaces handcrafted
fitness functions with LLM judges, a complementary direction to our work, which
retains the simulation-based fitness but diversifies the \emph{generator}.
\citet{flageat2022benchmarkingqd} provide a systematic benchmarking study of QD
algorithms on neuroevolution tasks, establishing baselines we draw on for
experimental protocol.
\citet{lehman2011novelty} established that novelty alone (without explicit quality
pressure) can drive exploration; our adversarial gossip mechanism adds quality
pressure from cross-model opponents.

\paragraph{Distributed and island-model EAs:}
The island model~\citep{cantupaz2001parallel} is the classical blueprint for
distributing evolutionary search: subpopulations evolve in parallel and exchange
migrants periodically. \citet{castillo2011distributed} extends this to REST-based
web services. Our asynchronous gossip substrate is in the spirit of the island model
but differs in three key ways: (i) migration uses publish-subscribe rather than
point-to-point topology, (ii) compute budgets across nodes are heterogeneous by
design, and (iii) the \emph{identity} of the LLM rather than random drift is the
primary source of diversity. Core War has been explored with island-model
co-evolution previously~\citep{corno2003corewarisland}, but without LLM operators
or QD objectives.

\paragraph{Adversarial co-evolution and Red Queen dynamics:}
The Red Queen hypothesis~\citep{vanvalen1973newlaw} describes evolutionary
arms races in which species must continually adapt simply to maintain relative
fitness against co-evolving competitors. \citet{hillis1990coevolving} brought
this dynamic into computational search by co-evolving sorting networks against
adversarial test cases, demonstrating that competitive co-evolution can escape
local optima where direct optimisation stalls.
\citet{rosin1997competitive} formalised the methodology for competitive
co-evolution in evolutionary algorithms, introducing hall-of-fame archives and
shared sampling to stabilise the arms race. \citet{kumar2026digitalredqueen}
extended this lineage to LLM-driven program evolution with the Digital Red
Queen framework, using Core War as a testbed and instantiating Red Queen
pressure inside MAP-Elites by retaining each round's champion as an opponent
in subsequent rounds. We extend this further to the distributed setting:
cross-node gossip delivers champions from heterogeneous LLMs as opponents,
creating inter-model Red Queen pressure that cannot arise in single-model
self-play.

\paragraph{Code generation with LLMs:}
Our system relies on the ability of diverse LLMs, Codex-family models
\citep{chen2021codex}, instruction-tuned models like Claude and Gemini, and
open-weight models such as Qwen, to generate syntactically valid Redcode warriors.
AlphaCode~\citep{li2022alphacode} demonstrated competition-level program synthesis;
our work repurposes diverse code-generation capabilities as QD search operators.

\paragraph{Diversity aware RL and multi-agent reasoning:}
There is growing evidence in reinforcement learning that diversity can play a strong role in reasoning
and multi-agent systems. DARLING~\citep{li2025darling} jointly optimises a
quality reward with a learned semantic-diversity signal during online RL post-training,
and reports gains in both quality and novelty over quality-only RL on instruction-following
and creative-writing benchmarks, alongside improved pass@$k$ on competition math.
DIVER~\citep{hu2025diver} adds a global, sequence-level diversity intrinsic reward via
potential-based reward shaping inside the RL-with-verifiable-rewards paradigm, and beats
competing RLVR baselines on in- and out-of-domain reasoning at both pass@$1$ and pass@$k$.
At inference time, multi-agent debate~\citep{liang2024encouraging} elicits divergent
thinking by having several LLM agents argue under a judge, addressing the degeneration
of thought failure mode of single-agent self-reflection.
These papers compliment this work and and inspire our exploration of diversity applied to
distributed Quality-Diversity search.

\section{Conclusion}
\label{sec:conclusion}

We introduced \method{}, a distributed Quality-Diversity search framework that
exploits the distinct inductive biases of heterogeneous LLMs as the primary source
of behavioral diversity. By connecting nodes via an asynchronous distributed
communication layer, we enable cross-model adversarial pressure and niche novelty without
synchronization barriers. A coverage argument shows that heterogeneous BC
distributions yield strictly higher expected archive coverage than any single
distribution at equal compute, and our experiments confirm this empirically:
at equal total LLM call budget, the merged diverse archive achieves $+124\%$
higher QD-Score and $+28\%$ more coverage than a solo node. We argue that
\emph{model diversity}, not just parallelism, should be a first-class design
principle in distributed LLM-based search.

\section*{Broader Impact}
\label{sec:impact}

This work empirically advances distributed evolutionary search using heterogeneous LLM ensembles.
The asynchronous, gossip-based architecture enables researchers without dedicated
GPU clusters to contribute to collaborative search by running local open-weight
models alongside cloud-hosted counterparts. Democratizing participation in
LLM-driven program synthesis could accelerate scientific discovery in domains
ranging from combinatorial optimization to automated theorem proving.
The Quality-Diversity framing also produces richer, more interpretable result
sets than single-objective optimization, supporting human-in-the-loop workflows
where practitioners inspect the full archive rather than a single champion.

\bibliographystyle{plainnat}
\bibliography{references}

\appendix

\section{MARS Configuration Details}
\label{app:mars}

All simulations use the following MARS configuration, held constant across all
experimental conditions:

\begin{itemize}
  \item Core size: 8{,}000 instructions
  \item Maximum cycles per battle: 80{,}000
  \item Rounds per pair: 20
  \item Initial warrior placement: random, minimum separation enforced
  \item Process limit per warrior: unlimited (standard MARS)
\end{itemize}

\section{LLM Prompt Templates}
\label{app:prompts}

Two prompt templates govern LLM calls:

\textbf{New warrior prompt} (used for 10\% of calls): instructs the LLM to generate
a novel Redcode warrior from scratch, given a description of Core War rules and the
current archive state.

\textbf{Mutation prompt} (used for 90\% of calls): provides an existing warrior as
context along with its fitness score and BC coordinates, and asks the LLM to produce
an improved variant.

Full prompt text is included in the code release and in the DRQ repository: \url{https://github.com/SakanaAI/drq/}.

\section{Network Implementation Details}
\label{app:implementation}

\subsection{Yggdrasil Overlay}
Yggdrasil~\citep{arceliar2019yggdrasil} assigns each node a stable IPv6 address derived from its public
key and performs NAT traversal via a distributed spanning-tree routing scheme. This allows nodes behind
firewalls or consumer routers to participate without manual port forwarding.

\subsection{AXL: Application Interface to the Network Layer}
\label{app:axl}

The bridge between the DRQ application and the Yggdrasil transport is
provided by AXL~\citep{gensyn2026axl}, an open-source P2P network for decentralized
agentic and AI/ML applications, developed by Gensyn AI. AXL is a Go binary that
embeds the Yggdrasil core and exposes a
local HTTP API for application interface.  The DRQ process communicates
exclusively with this local interface and never opens sockets to remote peers
directly.

\paragraph{Architecture.}
AXL consists of four layers: (i)~an \emph{HTTP API} that accepts application
requests on localhost; (ii)~an inbound \emph{message multiplexer} that
dispatches arriving TCP messages to the appropriate handler; (iii)~a \emph{userspace TCP/IP stack} (gVisor) that requires
no TUN device or root privileges; and (iv)~the \emph{Yggdrasil core}, which
manages the node's ed25519 keypair, derives a stable \texttt{200::/7} IPv6
address from the public key, and peers over TLS/TCP.

\paragraph{API.}
The three endpoints used by DEI are:
\begin{itemize}
  \item \texttt{POST /send} --- fire-and-forget delivery of a raw byte
    payload to a peer identified by its 64-character hex-encoded ed25519 public
    key (header \texttt{X-Destination-Peer-Id}).
  \item \texttt{GET /recv} --- polling dequeue of inbound messages; returns
    \texttt{204} when the queue is empty, or \texttt{200} with the raw payload
    and an \texttt{X-From-Peer-Id} header identifying the sender.
  \item \texttt{GET /topology} --- returns this node's IPv6 address, public
    key, and current peer list, used at startup to obtain the node identity
    shared with collaborating nodes.
\end{itemize}

\subsection{GossipSub Protocol}
We implement Gossipsub on top of AXL over Yggdrasil 
as the asynchronous communication layer for our ensemble DRQ experiments.  
GossipSub~\citep{vyzovitis2020gossipsub} maintains a \emph{mesh} of $D$ peers per topic
to which it eagerly pushes full message payloads. We use $D = 3$, giving a propagation 
diameter of $O(\log N)$ hops for $N$ nodes. Beyond the mesh, the node lazily announces 
message availability via \texttt{IHAVE} control messages; recipients lacking the announced 
message may request it with \texttt{IWANT}. A periodic heartbeat (default 1\,s) triggers graft 
and prune operations to repair the mesh after churn.

\section{LLM and Agent Usage}
\label{app:llm_usage}

In the spirit of disclosure, we used an LLM-based coding and writing agent in
the preparation of this paper. Specifically, the agent was used to:
(i)~co-author and revise prose in the manuscript, including drafting,
restructuring, and tightening sections under the authors' direction;
(ii)~debug the experimental codebase; and (iii)~write the plotting
code that generates the tables and figures presented in the paper. All
scientific claims, experimental design choices, and final wording were
reviewed and approved by the authors, who take responsibility for the
content.

\end{document}